\documentclass[review]{elsarticle}

\usepackage{lineno,hyperref}
\modulolinenumbers[5]
\pdfoutput=1
\journal{Cognitive Systems Research}

\begin{document}

\begin{frontmatter}

\title{Using an ensemble color space model to tackle adversarial examples\tnoteref{mytitlenote}}
\tnotetext[mytitlenote]{Supported by NSFC project Grant No. U1833101, Shenzhen Science and Technologies project under Grant No. JCYJ20160428182137473 and the Joint Research Center of Tencent and Tsinghua.}

\author[mymainaddress]{Shreyank N Gowda \corref{mycorrespondingauthor}}
\cortext[mycorrespondingauthor]{Corresponding author}
\ead{s1960707@ed.ac.uk}

\author[mysecondaryaddress]{Chun Yuan}
\ead{yuanc@sz.tsinghua.edu.cn}

\address[mymainaddress]{The University of Edinburgh, UK}
\address[mysecondaryaddress]{Tsinghua University, China}

\begin{abstract}
Minute pixel changes in an image drastically change the prediction that the deep learning model makes. One of the most
significant problems that could arise due to this, for instance,
is autonomous driving. Many methods have been proposed to
combat this with varying amounts of success. We propose a
3 step method for defending such attacks. First, we denoise
the image using statistical methods. Second, we show that
adopting multiple color spaces in the same model can help us
to fight these adversarial attacks further as each color space
detects certain features explicit to itself. Finally, the feature
maps generated are enlarged and sent back as an input to obtain even smaller features. We show that the proposed model
does not need to be trained to defend an particular type of attack and is inherently more robust to black-box, white-box,
and grey-box adversarial attack techniques. In particular, the
model is 56.12 percent more robust than compared models in
case of white box attacks when the models are not subject to
adversarial example training.
\end{abstract}


\end{frontmatter}


\section{Introduction}

Deep learning models have achieved extraordinary results in computer vision and natural language processing
tasks due to the extensive amount of computing
power available. They have shown near human capabilities in speech recognition\cite{hinton2012deep}, natural language processing\cite{collobert2008unified}, scene understanding tasks\cite{karpathy2015deep} and video recognition
\cite{gowda2017human}. Although these algorithms show great promise, neural networks are black-box systems because there is restricted
theoretical comprehension for their behavior. This lack
of understanding has the potential to be dangerous. 

Recent
research has shown them to be low on security. Small perturbations to existing data have shown to yield negative classification results
with high confidence. Recent research\cite{goodfellow2014explaining}\cite{kurakin2016adversarial}\cite{szegedy2013intriguing} has shown how these adversarial examples can be generated off existing data. These deftly crafted adversarial
examples can also force the neural network to classify the given
image into any chosen target class. These vulnerabilities of neural networks escalate critical
questions regarding the robustness, dependability, and security of deep learning applications. Figure 1 shows one such example.

\begin{figure}[h]
\begin{center}
   \includegraphics[width=1.0\linewidth]{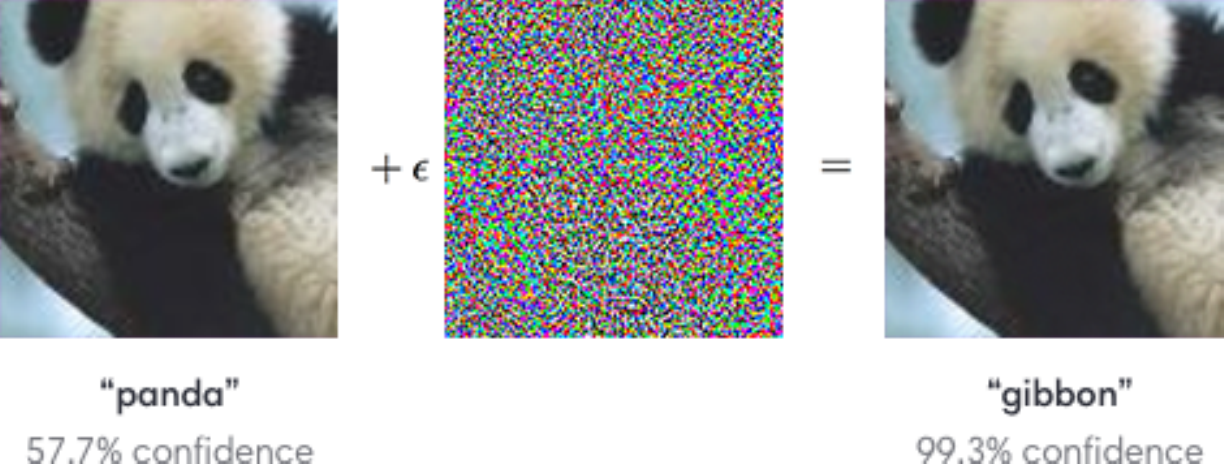}
\end{center}
   \caption{An example of how noise added can change classification result of a model [5] (in this case from a panda to a gibbon)}
\label{fig:cm}
\end{figure}

In \cite{carlini2016hidden} and \cite{nguyen2015deep} it was seen that
speech or image input can be modified to resemble relevant
input to the neural network but look/sound like complete
gibberish to humans. In \cite{gowda2016fiducial}, real-time input captured through a cell phone camera
can be subject to adversarial perturbations that result in complete mis-classification.  Authors in
\cite{feinman2017detecting} proposed the use of manifold distance, while, \cite{metzen2017detecting} utilized principal components to recognize adversarial attacks. These can easily be skipped through by prior
knowledge about the nature of the system. 
Optimization-based methods that are utilized to develop
provably-secure models are presently restricted to relatively
smaller networks and the guarantees provided are insufficient for real-life applications \cite{madry2017towards}\cite{raghunathan2018certified}.

Our paper provides a novel idea to make a model more robust without having the need to train the model separately for attacks. 
Following are the steps,
\begin{itemize}
    \item First, we perform statistical denoising of image. Second, we use an ensemble model consisting of
small but wide Densenet \cite{huang2017densely} models. 
    \item Each model will be
having an input mutated into a different color space.  
    \item Third,
the feature maps generated are enlarged in size and sent
again to the start of each densenet, to retrain the model to
obtain even smaller and precise features.  

\end{itemize}
The details of the model
and the reasons behind our choices will be explained in
further sections.

\section{Related Work}

We provide a compendium of adversarial attacks and techniques to defend against them in the following subsections. The input image can be considered as a 3-D tensor \textbf{I} where the three dimensions are height, width, and colors. Height and width represent the dimensions of the image and color represents the number of colors the image displays. 

Adversarial examples can be broadly classified to two types: targeted and untargeted. Targeted examples implies that the generated example compels the network to output any given image into a particular class and an untargeted example simply makes the network mis-classify an image. 
Every example involves adding a distortion \(\delta\) to the image. \textit{L\begin{footnotesize}
p
\end{footnotesize}} norms are often used as measures for these distortions. 

\subsection{Adversarial Attacks}

Adversarial samples were originally described in \cite{szegedy2013intriguing}. This subsection will briefly look into some of these attacks and what these attacks do. Fast gradient sign method (FGSM) was described in \cite{goodfellow2014explaining}. It is an \textit{L\begin{footnotesize}
\(\infty\)
\end{footnotesize}} type of attack and uses the gradient obtained from the loss function to determine the direction of modification for the pixels. It is represented in (1). \(\epsilon\) represents the magnitude of distortion added and y is the true label. 

\begin{equation}
\mathbf{I'}=\mathbf{I}-\epsilon .sign(\bigtriangledown_\mathbf{I}J(\mathbf{I},y_{true}) )
\end{equation}

The randomized fast gradient sign method (RAND-FGSM) was proposed in \cite{tramer2017ensemble}. The mathematical explanation is shown in (2) and (3).

\begin{equation}
\mathbf{I'}=\mathbf{I}+rand(\alpha)
\end{equation}
\begin{equation}
\mathbf{I''}=\mathbf{I'}+(\epsilon-\alpha) .sign(\bigtriangledown_\mathbf{I'}J(\mathbf{I'},y_{true}) )
\end{equation}

The Carlini-Wagner attack proposed in \cite{carlini2017towards} is an extremely powerful optimized based adversarial attack. It is a series of \textit{L\begin{footnotesize}
o
\end{footnotesize}}, \textit{L\begin{footnotesize}
2
\end{footnotesize}} and \textit{L\begin{footnotesize}
\(\infty\)
\end{footnotesize}} attacks with comparatively lower distortions with the above mentioned attacks. The value of constant c $>$ 0 and steps to obtain it can be found in \cite{carlini2017towards}.


Another attack method using FGSM as base was shown in \cite{kurakin2016adversarial}. It is called basic iterative method. It applies FGSM multiple times with a smaller step size. Equation (4),(5) describes it.
\begin{equation}
\mathbf{I'_{0}}=\mathbf{I}
\end{equation}
\begin{equation}
\mathbf{I'_{n+1}}=\mathbf{I'_{n}}+\alpha .sign(\bigtriangledown_\mathbf{I}J(\mathbf{I'_{n+1}},y_{true}) )
\end{equation}

Another form of attack is called the Jacobian saliency attack\cite{papernot2016limitations}. It is a computationally expensive algorithm. It is an \textit{L\begin{footnotesize}
o
\end{footnotesize}} form of attack. It basically chooses the most influential pixels by utilizing a greedy algorithm that calculates the Jacobian saliency map. Fault injection attack was proposed in \cite{liu2017fault}. The idea was to make slight changes to the DNN parameters to cause misclassification. 

An attack called Deepfool (DF) was proposed in \cite{moosavi2016deepfool}. It determines an
approximation of the decision boundary to perceive possible locations for adversarial perturbations. Attacks performed knowing the classifier details such as model parameters, weights etc are called white box attacks. Some attacks treat the classifier as a black box and hence these attacks are called black-box attacks.

\subsection{Defense strategies}

One of the earliest proposed techniques was to train the neural network to adversarially generated samples \cite{tramer2017ensemble}. Training with adversarial data made the neural network more robust to adversarial attacks. Label-smoothing \cite{warde201611} was shown to perform well against certain types of adversarial attacks. In the context of purifying the samples, generative models have been utilized \cite{oord2016pixel} to varying degrees of success. 

Use of auxiliary networks to clean samples also showed marked improvement in results against adversarial attacks\cite{oord2016pixel}\cite{meng2017magnet}. In \cite{papernot2016distillation} defensive distillation was proposed, it trains the classifier using a variant of distillation \cite{hinton2015distilling}.  While it gave improved results for white-box attacks, it failed to protect the network from black-box attacks \cite{carlini2017towards}. 

MagNet \cite{meng2017magnet} trains a collection of autoencoder networks (called as reformer networks) to make adversarial examples seem more natural. At test time, one autoencoder is randomly chosen and this magnifies the strength of their strategy. More recently, Defense-GAN was proposed \cite{samangouei2018defense}, it uses the generator part of the GAN to sanitize the input before passing it to the classifier. APE-GAN \cite{jin2019ape} amended the generator part to clean the adversarial samples and the discriminator differentiates between real and adversarial inputs.

\section{Proposed Method}

\subsection{Denoising and Color spaces}

Denoising can be considered as a minimizing problem where we minimize a loss and a penalty. We propose the use of the L1-norm of pixel updates as the penalty. The advantage of using L1-norms is that we deal directly with only the noisy pixels. We are given a noisy wavelet coefficient of a noisy image y$_{i}$, we want to recover the noise-free wavelet coefficient x$_{i}$ via a MAP estimator $\hat{x}_{i}$. $\hat{x}_{i}$. We assume that the noise in the image can be represented in the form of Gaussian noise with zero mean and standard deviation $\sigma_{n}$. Then we can express $p(y_{i}|x_{i})$ using Baye's rule as seen in (6). The advantage of using this non-parametric approach is that the model automatically adapts to the observed image data. 

\begin{equation}
p(y_{i}|x_{i}) = \frac{1}{{\sigma_{n} \sqrt {2\pi } }}e^{{{ - \left( {y_{i}- x_{i}} \right)^2 } \mathord{\left/ {\vphantom {{ - \left( {x - \mu } \right)^2 } {2\sigma_{n} ^2 }}} \right. \kern-\nulldelimiterspace} {2\sigma_{n} ^2 }}}
\end{equation}

A color space is basically an organization of colors. We can also look at color spaces as an abstract mathematical
model that aids us to portray colors as numbers.
For more details about the theory between different color spaces, one can refer to \cite{rasouli2017effect}. ColorNet \cite{gowda2018colornet} shows us that using an ensemble model of different color spaces help us to obtain state of the art classification results. Also, based on the confustion matrix in \cite{gowda2018colornet} we can say that the activation map for each image is slightly different after preprocessing the image into a different color space.

\subsection{Architecture of model and novelties proposed}

We first denoise the images as explained earlier. Next, we pass the image after denoising to our network. Our network comprises of 4 small but wide DenseNets\cite{zagoruyko2016wide}. We use a wide-approach for the Densenet \cite{huang2017densely} as it reduces the parameters. The denoised image is sent as input to the first densenet, while further preprocessing is done in the form of color space conversion to LAB, HSV and YUV. We reduce the number of color spaces based on individual accuracy \cite{gowda2018colornet}\cite{gowda2020stegcolnet} to aid the time of execution needed for our approach. The final activation map is taken and passed as a recurrent input after resizing the image based on the activation map size. This process is shown in Figure 2. This progressive resizing helps to detect tinier features, increasing the robustness of the network. The entire architecture is shown in Figure 3.The entire architecture is shown in Figure 3. We take the feature map at the last layer, collect the part of the image with the highest density and enlarge the image to the original size of the image (for instance 32x32 in the CIFAR datasets \cite{krizhevsky2009learning}). Hence, each image is used as an input twice to the model.

\begin{figure}[h]
\begin{center}
   \includegraphics[width=1.0\linewidth]{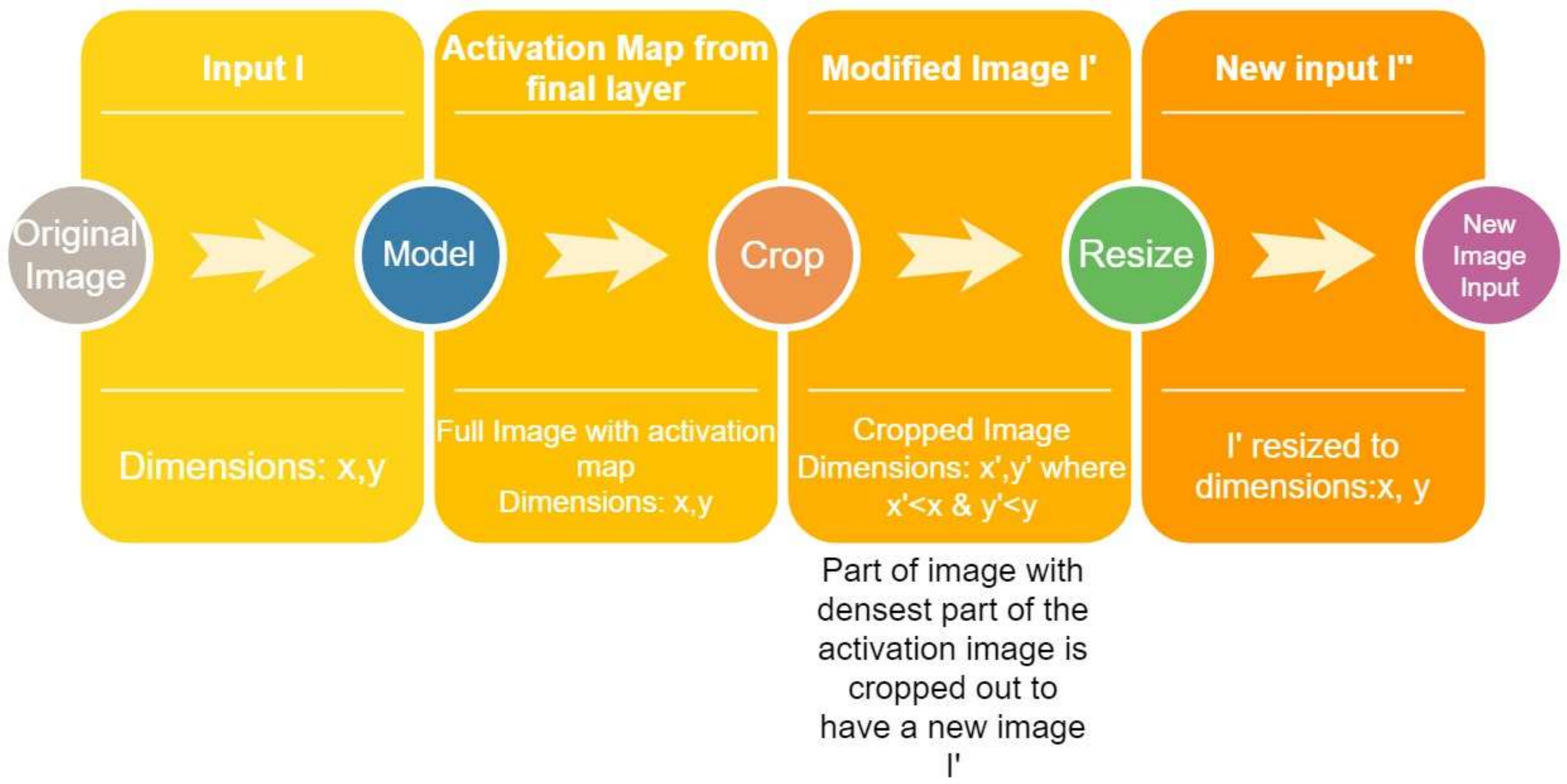}
\end{center}
   \caption{Progressive resizing of activation map flowchart}
\label{fig:cm}
\end{figure}

\begin{figure}[h]
\begin{center}
   \includegraphics[width=1.0\linewidth]{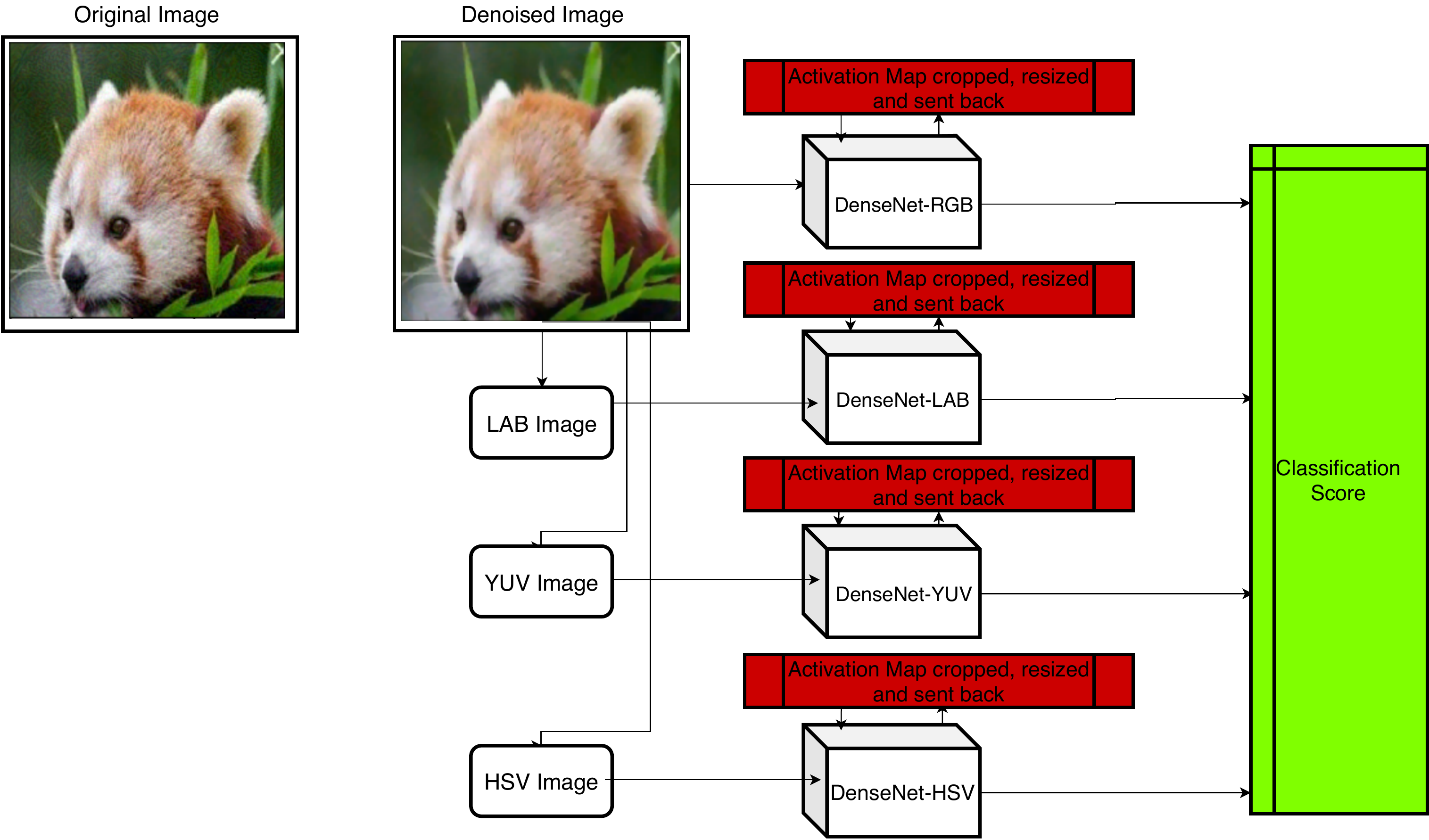}
\end{center}
   \caption{Architecture of proposed model}
\label{fig:cm}
\end{figure}

First, we denoise the image using a statistical approach. Second, to the best of our knowledge, color spaces have not been explored with regards to adversarial attacks. Third, progressive resizing, is a novel technique that uses the feature map of an image to augment the dataset by enlarging the map and using the same as an input to the model.

\section{Experimental Analysis}

We test our model on the 3 types of attacks discussed earlier namely: white-box, grey-box and black-box attacks.

\subsection{Training}

As discussed in the earlier section, our proposed model denoises the original input image. This modified image is sent as input to one small densenet model. The image is simultaneously processed into LAB, YUV and HSV images and sent as inputs to corresponding densenets. The activation map at the final layer is sent back as an input to the densenet by cropping out the original input with the densest parts of the activation map, and this cropped image is resized to the original image and again sent as input. This progressive resizing technique is done to obtain lower level features that help to detect an object with higher accuracy.

\subsection{White-box attacks}

We compare our results with Resnet-50 \cite{he2015delving}, VGG-19 \cite{simonyan2014very}, Pixel-Deflection (PD) \cite{prakash2018deflecting}, MagNet \cite{meng2017magnet}. The algorithms used for the attacks are FGSM, IGSM, DFool, JSMA, C and W. Table 1 contains the comparison results. The results obtained are from choosing a random subset of the imagenet dataset and taking the average prediction over 6 runs. Since, the training involves addition of a different adversarial noise we get different results. Averaging them gives a higher level of accuracy. For reference purposes, table 2 highlights the results of how the various models perform when trained to deal with particular types of adversarial examples. As can be seen from table 1 our proposed model is significantly more robust to adversarial attacks and performs reasonably well without training the model for defense against attacks. In terms of numbers, the proposed method is \textbf{56.12 percent} more robust without training. Although, after training it performs slightly worse than the other mentioned models. Even then, the model only performs 1.08 percent worse on average.

\setlength{\tabcolsep}{1.4pt}

\setlength{\tabcolsep}{4pt}
\begin{table}[t]
\caption{Classification results without training to defend}
\vskip 0.15in
\begin{center}
\begin{small}
\begin{sc}
\begin{tabular}{llllll}
\hline
Attack & L2 & Resnet & VGG-19 & P-D & \textbf{Proposed}\\
\hline
FGSM  & 0.05 & 20.3 & 12.4 & 20.1 & \textbf{78.3}\\
DFool  & 0.03 & 25.8 & 23.7 & 26.3 & \textbf{72.8}\\
IGSM  & 0.04 & 14.3 & 9.7 & 14.1 & \textbf{79.6}\\
C and W  & 0.02 & 4.1 & 0.00 & 4.7 & \textbf{71.6}\\
JSMA  & 0.03 & 25.5 & 29 & 25.5 & \textbf{72.9}\\
\hline
\end{tabular}
\end{sc}
\end{small}
\end{center}
\vskip -0.1in
\end{table}
\setlength{\tabcolsep}{1.4pt}

\setlength{\tabcolsep}{1.4pt}

\setlength{\tabcolsep}{4pt}
\begin{table}[t]
\caption{Classification results after training to defend}
\vskip 0.15in
\begin{center}
\begin{small}
\begin{sc}
\begin{tabular}{llllll}
\hline
Attack & L2 & Resnet & VGG-19 & P-D & \textbf{Proposed}\\
\hline

FGSM  & 0.05 & 79.2 & 79.0 & 79.9 & \textbf{81.4}\\
DFool  & 0.03 & \textbf{86.3} & 84.2 & 86.2 & 84.3\\
IGSM  & 0.04 & \textbf{83.9} & 78.9 & \textbf{83.9} & 83.7\\
C and W  & 0.02 & 92.9 & \textbf{93.1} & 92.9 & 90.9\\
JSMA  & 0.03 & 91.3 & \textbf{93.3} & 91.6 & 90.8\\

\hline
\end{tabular}
\end{sc}
\end{small}
\end{center}
\vskip -0.1in
\end{table}
\setlength{\tabcolsep}{1.4pt}

\subsection{Black-box and grey-box attacks}

Generating black box attacks are more difficult than white-box attacks and hence we use models that already have black-box attacks performed on them and compare the results with our model. Again, we use the vast imagenet dataset \cite{deng2009imagenet}. We compare the top-1 accuracy of our model against the models proposed in \cite{moosavi2018divide} (divide, denoise and dispatch or D3) and \cite{guo2017countering}. Table 3 shows the results obtained. Attacks used were deep fool and FGSM. In this case, it can be seen that the proposed model performs significantly better than the models in comparison and that the black-box attacks have insignificant impact on the model.

\setlength{\tabcolsep}{4pt}
\begin{table}[t]
\caption{ Comparison of top-1 accuracy on imagenet against blackbox attacks}
\vskip 0.15in
\begin{center}
\begin{small}
\begin{sc}
\begin{tabular}{llll}
\hline
Network & No-attack & DFool & FGSM\\
\hline
D3 (n = 40K, k = 5)  & 71.8 & 63.1 & 68.3\\
D3 (n = 10K, k = 5)   & 70.8 & 64.6 & 68.5\\
Quilting [42]  & 70.1 & 65.2 & 65.5\\
TVM + QUILTING [42]  & 72.4 & 65.8 & 65.7\\
\textbf{Proposed Method}  & \textbf{78.3} & \textbf{76.9} & \textbf{77.4}\\

\hline
\end{tabular}
\end{sc}
\end{small}
\end{center}
\vskip -0.1in
\end{table}
\setlength{\tabcolsep}{1.4pt}

Grey-box attacks are attacks wherein the adversary has access to the network weights, but does not have an idea about the defense mechanism in place. We do the same comparisons as with the black-box attacks. Table 4 shows the results obtained. Again, as seen with black-box attacks, the model shows high robustness to adversarial attacks.

\setlength{\tabcolsep}{4pt}
\begin{table}[t]
\caption{ Comparison of top-1 accuracy on imagenet against grey-box attacks}
\vskip 0.15in
\begin{center}
\begin{small}
\begin{sc}
\begin{tabular}{llll}
\hline
Network & No-attack & DFool & FGSM\\
\hline

D3 (n = 40K, k = 5)  & 71.8 & 57.8 & 67.3\\
D3 (n = 10K, k = 5)   & 70.8 & 62.3 & 67.5\\
Quilting [42]  & 69.7 & 34.5 & 39.8\\
TVM + QUILTING [42]  & 66.3 & 44.7 & 31.4\\
\textbf{Proposed Method}  & \textbf{79.6} & \textbf{78.2} & \textbf{81.3}\\

\hline
\end{tabular}
\end{sc}
\end{small}
\end{center}
\vskip -0.1in
\end{table}
\setlength{\tabcolsep}{1.4pt}

\section{Conclusion}

In this paper, we propose a novel algorithm that shows robustness to adversarial attacks. It follows a 3 step process. First, the image is preprocessed by denoising all the images. This helps the model to not overfit on specific features. Next, we use a color-space ensemble model. This was on the basis that certain classes of images were being represented better by certain color spaces. Finally, we use progressive image resizing by cropping out the part of the image that has highest density of features and then resizing into the original input size and passing it back to the network as an input. This helps us detect even smaller features and hence can detect an object correctly under adversarial conditions. Based on the results we could see that the proposed model was 56.12 percent more robust without training.

\bibliographystyle{elsarticle-num-names}
\bibliography{mybibfile.bib}

\end{document}